\documentclass[10pt,twocolumn,letterpaper]{article}
\usepackage{cvpr}              

\usepackage{xcolor}
\usepackage{dsfont}
\usepackage{amsmath,amsfonts}
\usepackage{algorithmic}
\usepackage{algorithm}
\usepackage{lscape}
\usepackage{array}
\usepackage{textcomp}
\usepackage{stfloats}
\usepackage{url}
\usepackage{float}
\usepackage{booktabs}
\usepackage{verbatim}
\usepackage{graphicx}
\usepackage{cite}
\usepackage{colortbl,hhline}
\usepackage{subcaption}
\usepackage{multirow}
\usepackage{multicol}
\usepackage{pifont}
\definecolor{cvprblue}{rgb}{0.21,0.49,0.74}

\usepackage[pagebackref,breaklinks,colorlinks,allcolors=cvprblue]{hyperref}

\DeclareRobustCommand{\IEEEauthorrefmark}[1]{\smash{\textsuperscript{\footnotesize #1}}}

\def\paperID{*****} 
\def\confName{CVPR}
\def\confYear{2026}

\title{MMLSv2: A Multimodal Dataset for Martian Landslide Detection in Remote Sensing Imagery}

\author{Sidike Paheding\IEEEauthorrefmark{1}, Abel Reyes-Angulo\IEEEauthorrefmark{2},  Leo Thomas Ramos\IEEEauthorrefmark{3,4}, 
    Angel D. Sappa\IEEEauthorrefmark{3,4,5}, \\  Rajaneesh A.\IEEEauthorrefmark{6}, Hiral P. B.\IEEEauthorrefmark{6},  Sajin Kumar K. S.\IEEEauthorrefmark{6},
    Thomas Oommen\IEEEauthorrefmark{7}\\
    \IEEEauthorrefmark{1}Fairfield University\quad
    \IEEEauthorrefmark{2}Michigan Technological University\quad
    \IEEEauthorrefmark{3}Computer Vision Center \\
    \IEEEauthorrefmark{4}Universitat Autònoma de Barcelona \quad
    \IEEEauthorrefmark{5}ESPOL Polytechnic University \quad
    \IEEEauthorrefmark{6}University of Kerala\\
    \IEEEauthorrefmark{7}University of Mississippi
}

\begin{document}
\maketitle

\begin{abstract}
We present MMLSv2, a dataset for landslide segmentation on Martian surfaces. MMLSv2 consists of multimodal imagery with seven bands: RGB, digital elevation model, slope, thermal inertia, and grayscale channels. MMLSv2 comprises 664 images distributed across training, validation, and test splits. In addition, an isolated test set of 276 images from a geographically disjoint region from the base dataset is released to evaluate spatial generalization. Experiments conducted with multiple segmentation models show that the dataset supports stable training and achieves competitive performance, while still posing challenges in fragmented, elongated, and small-scale landslide regions. Evaluation on the isolated test set leads to a noticeable performance drop, indicating increased difficulty and highlighting its value for assessing model robustness and generalization beyond standard in-distribution settings. Dataset will be available at: \url{https://github.com/MAIN-Lab/MMLS_v2}
\end{abstract}

\section{Introduction}

Landslides are mass movement processes in which soil, rock, or debris are displaced downslope under the influence of gravity \citep{9884949,10342762}, typically triggered by a combination of geological, morphological, and environmental factors \citep{lacroix_handwerger_2020}. While they play a significant role in shaping geomorphological landscapes \citep{GUZZETTI201242}, landslides are also among the most hazardous natural phenomena due to their sudden onset and potential for large-scale impact \citep{10640838,10342762}. As a result, landslide identification is essential for emergency response, land-use planning, and disaster risk mitigation \citep{9064562,wang_wang_dai_2024}.

\begin{figure*}[t]
    \centering
    \includegraphics[width=0.8\linewidth]{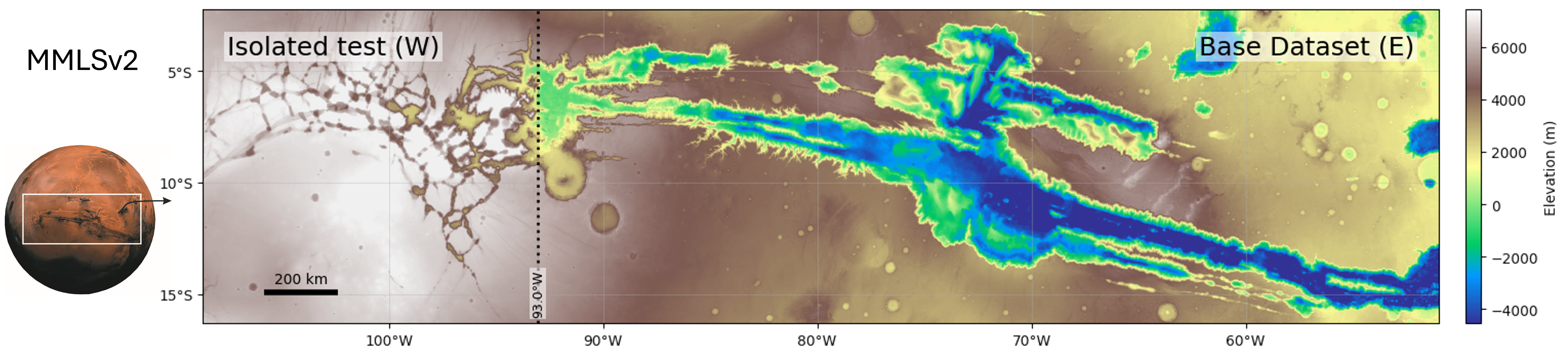}
    \caption{Location of Valles Marineris on Mars used in this study for constructing the MMLSv2 dataset.}
    \label{fig:valles_marineris}
\end{figure*}

Traditionally, landslide monitoring and recognition relied on field surveys or manual interpretation of satellite imagery \citep{9884949}, approaches that are time-consuming, difficult to scale \citep{9884949,AMATYA2021106000}, and poorly suited to the intrinsic complexity of landslides. Strong morphological variability in size, shape, spatial extent, and internal structure \citep{Zhong16022020} hinders consistent description and systematic analysis, motivating the increasing adoption of automated landslide identification methods based on artificial intelligence \citep{10641843}.

Specifically, deep learning (DL)-based semantic segmentation has become a standard formulation for landslide identification \citep{10246979}. These models are trained using large annotated datasets \citep{10623211,RSCVC_2025}, which allow them to learn the spatial organization, textural properties, and morphological patterns that distinguish landslide deposits from surrounding terrain \citep{kaushal_gupta_sehgal_2024}. This enables pixel-level delineation and more efficient analysis of landslide distributions than manual interpretation \citep{10246979,DU2021104860}.

Despite their effectiveness, DL methods for landslide segmentation are strongly constrained by data availability \citep{10641843,9323230}. Their performance depends on large, well-annotated datasets \citep{10641843,DU2021104860}, yet landslides are spatially sparse and rarely occur with sufficient density within a given region \citep{9323230}, resulting in severe class imbalance \citep{9323230,guptaaaa2020}. Moreover, generating reliable data remains challenging, particularly in regions with limited satellite coverage or expert-labelled annotations \citep{10641843}, representing a major bottleneck for robust and generalizable models.

Moreover, landslide identification is particularly challenging when relying solely on standard RGB imagery. Surface appearance varies widely across events, geological settings, and stages of evolution \citep{Zhong16022020,li_moon_2021}, and landslide deposits may be partially or fully obscured by vegetation, debris, or other land cover types \citep{Zhong16022020}, resulting in complex and non-negligible surface differences that hinder reliable identification \citep{Zhong16022020,DU2021104860} when relying solely on simple cues.

Building on the above, this work introduces the Multimodal Martian Landslide Dataset version 2 (MMLSv2) for binary semantic segmentation of landslides from orbital imagery. MMLSv2 adopts the Martian surface as a study scenario due to its predominantly rocky and arid conditions, which favour the clear expression and preservation of large-scale landslide structures, and because Mars is one of the planetary bodies most analogous to Earth in terms of surface processes and geomorphological dynamics \citep{10528357,farley_willhueso_2020}. The dataset incorporates seven complementary data channels: red, green, and blue spectral bands, digital elevation model, slope, thermal inertia, and grayscale representations, providing a richer characterization of surface properties for landslide analysis than RGB-only data.


We note that this work builds upon the MMLSv1 dataset introduced in \citep{10483716,reyes2023c,RAJANEESH2022114886}. The revised version incorporates substantial improvements, including refined and corrected pixel-level annotations, a dedicated data partitioning strategy, and more than 100 additional images. In addition, MMLSv2 introduces a fully isolated test set of 276 images from a geographically distinct region, enabling a more robust evaluation of model generalization. Together, these revisions provide a more reliable and challenging benchmark for landslide segmentation in planetary imagery. The main contributions of this work are summarized as follows:

\begin{itemize}
    \item Introduction of MMLSv2, a multimodal dataset for binary semantic segmentation of landslides from orbital imagery, composed of seven data channels.
    \item Inclusion of a fully isolated test set covering a geographically separate region, explicitly designed to support the evaluation of spatial generalization.
    \item Benchmark experiments demonstrating that MMLSv2 supports stable model convergence while preserving challenging conditions for landslide segmentation. 
   \item Public release of MMLSv2 to facilitate reproducible research and future developments.
\end{itemize}

\section{MMLSv2}

\subsection{Data acquisition}\label{sec:data_adq}


MMLSv2 focuses on the Valles Marineris region (Fig. \ref{fig:valles_marineris}), one of the largest canyon systems on Mars and a geomorphologically active environment characterized by steep escarpments, complex terrain, and pervasive mass-wasting processes. Owing to its diverse topographic conditions and extensive prior study in planetary geomorphology, this region provides a challenging and representative testbed for landslide analysis.

To support robust landslide mapping, we construct a multi-source dataset integrating thermal, optical, and topographic modalities. Thermophysical context is provided by nighttime infrared imagery from the Thermal Emission Imaging System (THEMIS) aboard Mars Odyssey \citep{Fergason}, together with the global thermal inertia mosaic at 100 m spatial resolution (USGS Astrogeology Science Center). Fine-scale geomorphology is captured using Context Camera (CTX) imagery from the Mars Reconnaissance Orbiter (MRO) \citep{Malin} at approximately 6 m resolution (Murray Lab CTX Portal), while topographic information is derived from Digital Elevation Models (DEMs) generated by the Mars Orbiter Laser Altimeter (MOLA) aboard Mars Global Surveyor (MGS) \citep{Smith}, blended with HRSC data at 200 m resolution (USGS Astrogeology Science Center), supporting slope computation, relief characterization, and assessment of landslide extent relative to regional terrain.

\subsubsection{Co-registration, landslide identification, and harmonization}

All data sources described previously originate from different missions and products, and therefore exhibit heterogeneous spatial resolutions, coverage extents, and native grids. To enable their joint use, all datasets are co-registered and processed in ESRI ArcGIS. 

Landslides were manually identified and digitized as polygons following established morphological criteria \citep{QUANTIN20041011,crosta2018introducing}, with study regions systematically labeled as landslide or non-landslide to produce a spatially explicit hazard inventory. Slope is derived from the MOLA DEM using the ArcGIS Slope tool, and a seven-band composite is constructed for methodological uniformity by integrating thermal inertia, slope, DEM, CTX imagery, RGB basemaps, and the Viking colorized global mosaic (232 m; USGS Astrogeology Science Center). This unified multimodal representation supports multi-scale geomorphological analysis, enabling qualitative characterization and quantitative assessment of landslide frequency, spatial distribution, and morphometric properties across Valles Marineris.

\subsubsection{Improvement made to the MMLSv1 dataset}

We refine the Valles Marineris landslide dataset described in \citep{10483716,reyes2023c,RAJANEESH2022114886}, termed MMLSv1, to improve label consistency and model generalization. In MMLSv1, annotations primarily captured idealized landslides comprising depletion, run-out, and depositional zones, leading to the exclusion of events lacking distinct depositional features. As a result, valid landslides were implicitly treated as negative samples, introducing label noise during training.

In CTX imagery, depletion and run-out regions exhibit distinctive geomorphological signatures, whereas depositional areas often share visual characteristics with other depositional environments, resulting in ambiguous supervision. Including such complex depositional regions can bias representation learning and degrade detection performance.

To address these issues, we augment the dataset with previously omitted landslides, including events without well-defined depositional zones, and remove complex depositional regions from existing annotations. This reduces spurious correlations and improves supervision quality, leading to more reliable learning and enhanced landslide mapping performance. The refined Multimodal Martian Landslide dataset is referred to as MMLSv2, where each input tile represents a spatially aligned multimodal observation with all channels resampled to a common resolution and pixel-level correspondence enforced across modalities.

\subsection{Mask alignment and raster-level quality control}

Although multimodal inputs are harmonized onto a common grid (Sec.~\ref{sec:data_adq}), the landslide inventory is produced independently and must be brought into exact pixel correspondence with the seven-band composite. The final mask GeoTIFF is therefore aligned to the reference composite by enforcing identical CRS, affine transform, width, and height, resampling with nearest-neighbor interpolation to preserve categorical labels. The mask is then binarized by mapping the landslide class (1) to foreground and all remaining values, including nodata, to background (0), yielding a clean $\{0,1\}$ target. As a sanity check, equality of CRS and affine transforms between the aligned mask and the composite is verified before any split or patch extraction, preventing silent sub-pixel shifts across modalities.

\subsection{Raster-level geographic holdout and patch extraction} 

To explicitly evaluate spatial generalization, we define a geographically disjoint holdout at the raster level by splitting the aligned Valles Marineris mosaics into two non-overlapping subregions using a fixed east-west boundary (Fig.~\ref{fig:valles_marineris}). All patches extracted from the holdout subregion form the isolated test set and are never used for training, or validation. Both the base region and the isolated region are tiled into fixed-size patches of $128\times128$ pixels using a deterministic grid with zero overlap. Patch origins are generated with a stride equal to the patch size, with an additional terminal origin to ensure boundary coverage without padding. Each patch is saved as a georeferenced GeoTIFF and indexed by its grid coordinates (\textit{col, row}).

\subsection{Partitioning strategy} \label{sec:part_strategy}

Before partitioning, we generate a deterministic grid of $128\times128$ patches from the base region (Sec.~\ref{sec:data_adq}). 
Partitioning is performed only on these base-region patches; the isolated test set is defined earlier via a raster-level geographic holdout and is kept fully separate from all training/validation decisions.

In patch based landslide segmentation, data partitioning is a critical yet often underestimated component of dataset design. Randomly assigning patches to training, validation, and test sets is ill suited to spatially continuous terrains, as neighbouring patches share highly correlated visual patterns. Distributing such patches across different splits introduces spatial information leakage, allowing the model to implicitly exploit spatial context during evaluation that is closely related to the training data, undermining the assumption of independence between subsets.

Also, random partitioning fails to capture the highly uneven distribution of landslide content across patches. Foreground coverage varies widely, from sparse occurrences to large continuous failure regions, and random splits do not guarantee that this variability is consistently represented. As a result, evaluation subsets may become unbalanced and weakly informative, biasing performance assessment toward memorization rather than true generalization, a critical issue in landslide segmentation where spatial continuity and generalization to unseen regions are essential.


Based on the considerations above, we adopt a dedicated partitioning strategy guided by three principles: (i) enforcing spatial independence between subsets, (ii) preserving a balanced distribution of sample difficulty across training, validation, and test sets, and (iii) ensuring reproducibility through deterministic and well-defined splitting criteria. This design provides a reliable and unbiased basis for model training and evaluation.


To achieve this, we combine foreground based stratification with spatial grouping to generate balanced and spatially coherent data splits. Partitioning is performed on spatial groups of patches rather than individual samples, jointly enforcing sample independence and difficulty balance.

Each patch is first characterized by its foreground ratio, defined as the proportion of landslide pixels in its binary segmentation mask. Given a mask $M \in \{0,1\}^{H\times W}$, the foreground ratio $r$ is computed as in Equation \ref{eq:1}:

\begin{equation} \label{eq:1}
    r = \frac{1}{HW}\sum_{i=1}^{H}\sum_{j=1}^{W}1(M_{ij}=1),
\end{equation}
where $H$ and $W$ denote the patch dimensions. This scalar provides a compact proxy for landslide dominance and sample difficulty.

To enforce spatial independence, patches are grouped according to spatial proximity using their grid indices (\textit{col, row}). A 2$\times$2 grouping scheme assigns each patch to a spatial block, as defined in Equation \ref{eq:2}:

\begin{equation}\label{eq:2}
    b = \begin{pmatrix} [\frac{\text{col}}{2}], [\frac{\text{row}}{2}]
\end{pmatrix}	.
\end{equation}

All patches within the same block are treated as an indivisible unit and assigned to the same split, preventing adjacent patches from being distributed across different subsets, while preserving sufficient spatial diversity.

Stratification is applied by computing, for each block, a representative foreground ratio as the average over its patches. Blocks are then stratified using quantile based binning to ensure consistent representation of landslide dominance across subsets. Partitioning follows predefined split proportions, with assignments propagated to all patches within each block. This strategy yields spatially independent, foreground balanced, and fully reproducible splits.

\section{Overview and dataset statistics}


Table \ref{tab:data_stats} summarizes the composition of MMLSv2 across the different splits. While the training, validation, and standard test sets exhibit comparable foreground statistics, the isolated test set shows a markedly lower average landslide coverage and reduced dispersion. This shows that the isolated split contains a larger proportion of sparse events, making it structurally different from the data used during training. Such a distribution is intentional and serves to assess the ability of models to generalize beyond the dominant patterns observed in the training data, rather than to interpolate within a similar foreground regime. The wide range of foreground ratios observed across all splits further confirms the heterogeneity of MMLSv2, ensuring exposure to both nearly empty patches and highly saturated landslide.


\begin{table}[t]
\caption{Distribution of the MMLSv2 dataset across the different splits. The foreground ratio is expressed as percentage of pixels belonging to landslide regions, including its average (Avg.$_{FG}$), standard deviation (Std.$_{FG}$) and minimum–maximum values (Min.$_{FG}$, Max.$_{FG}$).}\label{tab:data_stats}
\centering
\resizebox{\linewidth}{!}{%
\begin{tabular}{p{1.7cm}p{1.4cm}p{1.4cm}p{1.4cm}p{1.4cm}p{1.5cm}}
    \toprule
    \textbf{Split} & \textbf{\# Images} & \textbf{Avg.$_{FG}${\footnotesize (\%)}} & \textbf{Std.$_{FG}${\footnotesize (\%)}} & \textbf{Min.$_{FG}${\footnotesize (\%)}} & \textbf{Max.$_{FG}${\footnotesize (\%)}}\\
    \midrule
    Train & 465 & 35.41 & 25.64 & 0.02 & 99.52\\
    Val & 66 & 31.53 & 24.05 & 0.08 & 90.32\\
    Test & 133 & 33.82 & 25.05 & 0.10 & 90.67\\
    Isolated test & 276 & 21.83 & 17.08 & 0.01 & 71.95\\
    \bottomrule
\end{tabular}}
\end{table}

\begin{figure}[t]
    \centering
    \includegraphics[width=0.6\columnwidth]{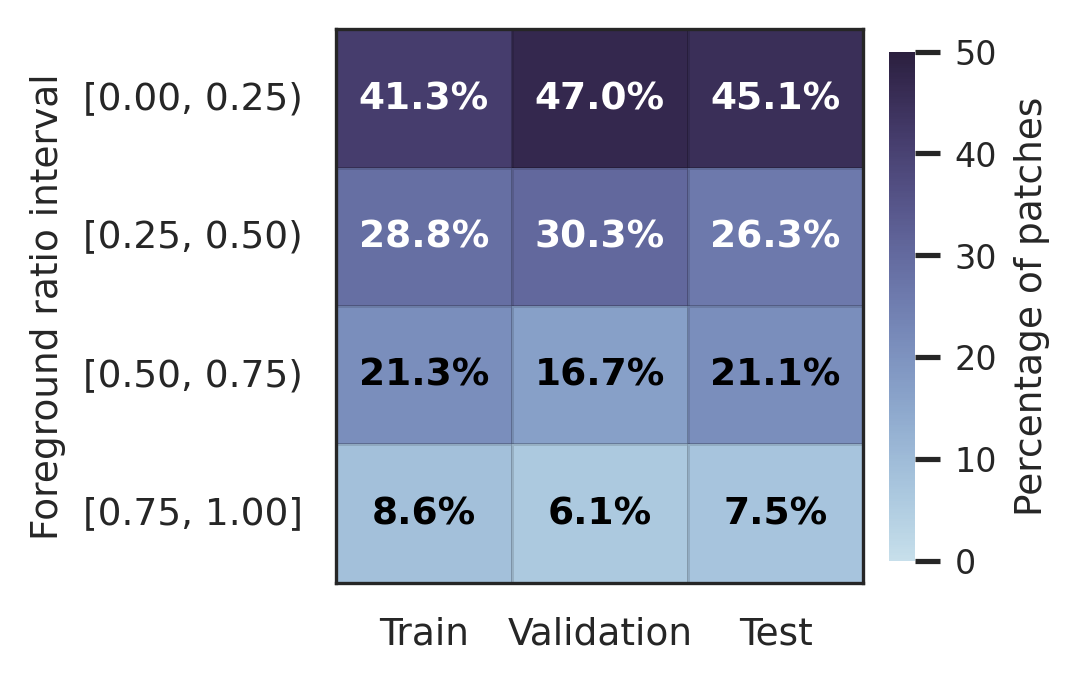}
    \caption{Distribution of foreground ratio intervals across the splits of MMLSv2. The three subsets exhibit highly consistent patterns, indicating that the proposed partitioning strategy preserves comparable levels of sample difficulty, with no split dominated by either nearly empty patches or highly foreground-dense samples.} \label{fig:bar_plots}
\end{figure}

Fig. \ref{fig:bar_plots} shows the distribution of foreground ratios across base dataset splits resulting from the proposed partitioning strategy. The consistent proportions of low, medium, and high foreground content indicate that no subset is biased toward sparse or highly saturated samples. This balance ensures that performance differences observed during evaluation are not driven by trivial variations in foreground density, but rather reflect the models’ ability to generalize across comparable levels of scene complexity.

\begin{figure*}[t]
    \centering
    \includegraphics[width=0.76\linewidth]{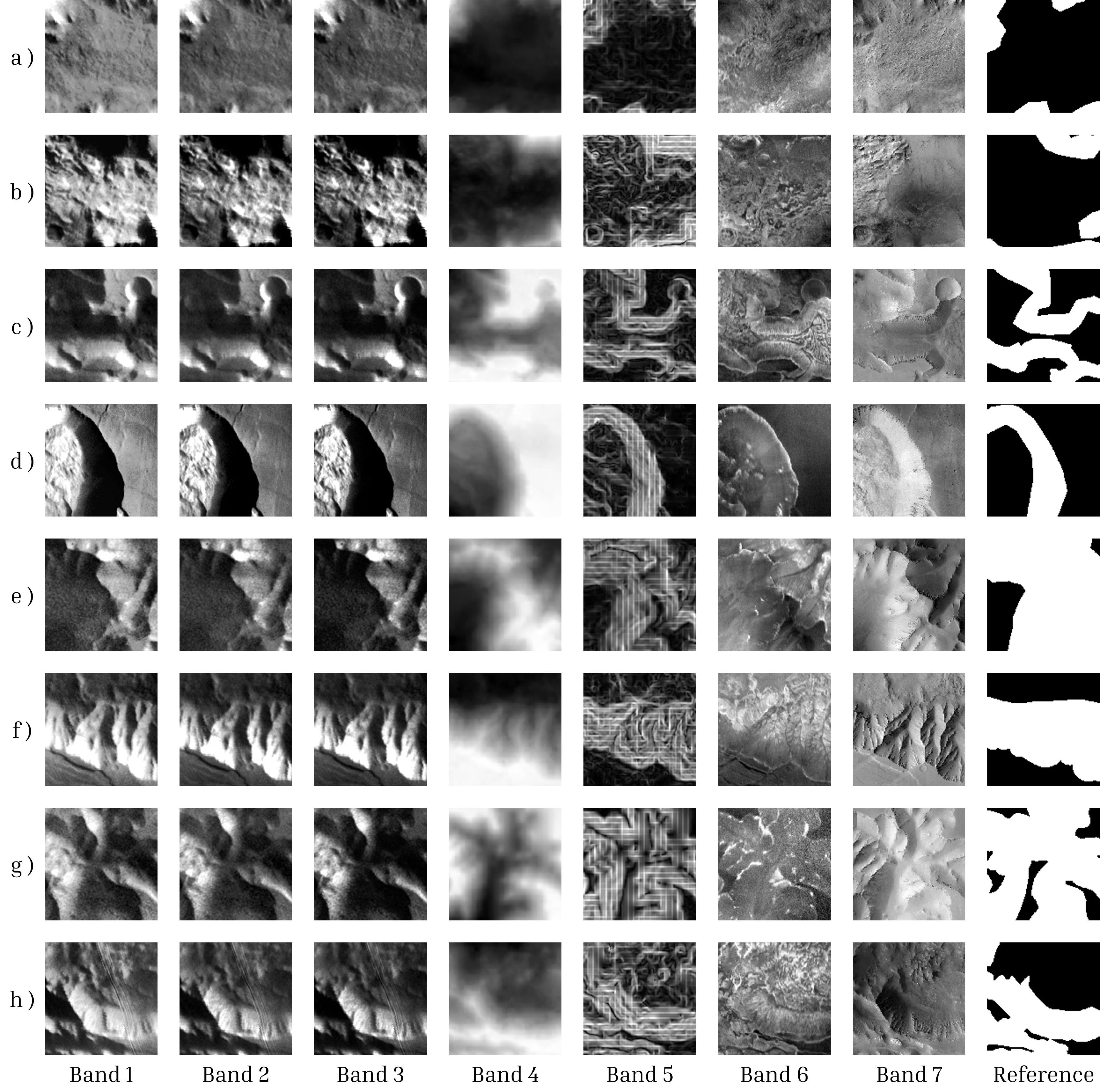}
    \caption{Representative scenes from the MMLSv2 dataset. Each row corresponds to a different image tile, while columns show the individual input bands composing the multimodal image, followed by the ground-truth landslide mask. The examples illustrate the wide morphological diversity captured by MMLSv2, including small, isolated events (rows a-b), elongated and curved landslides (rows c-d), extensive and continuous failure areas (rows e-f), and fragmented, irregular landslides occurring in complex geomorphological settings (rows g-h). Band order: (1) Red, (2) Green, (3) Blue, (4) DEM, (5) Slope, (6) Thermal inertia, (7) Grayscale.}
    \label{fig:mmlsv2_samples}
\end{figure*}

\begin{figure*}[t]
    \centering
    \includegraphics[width=0.76\linewidth]{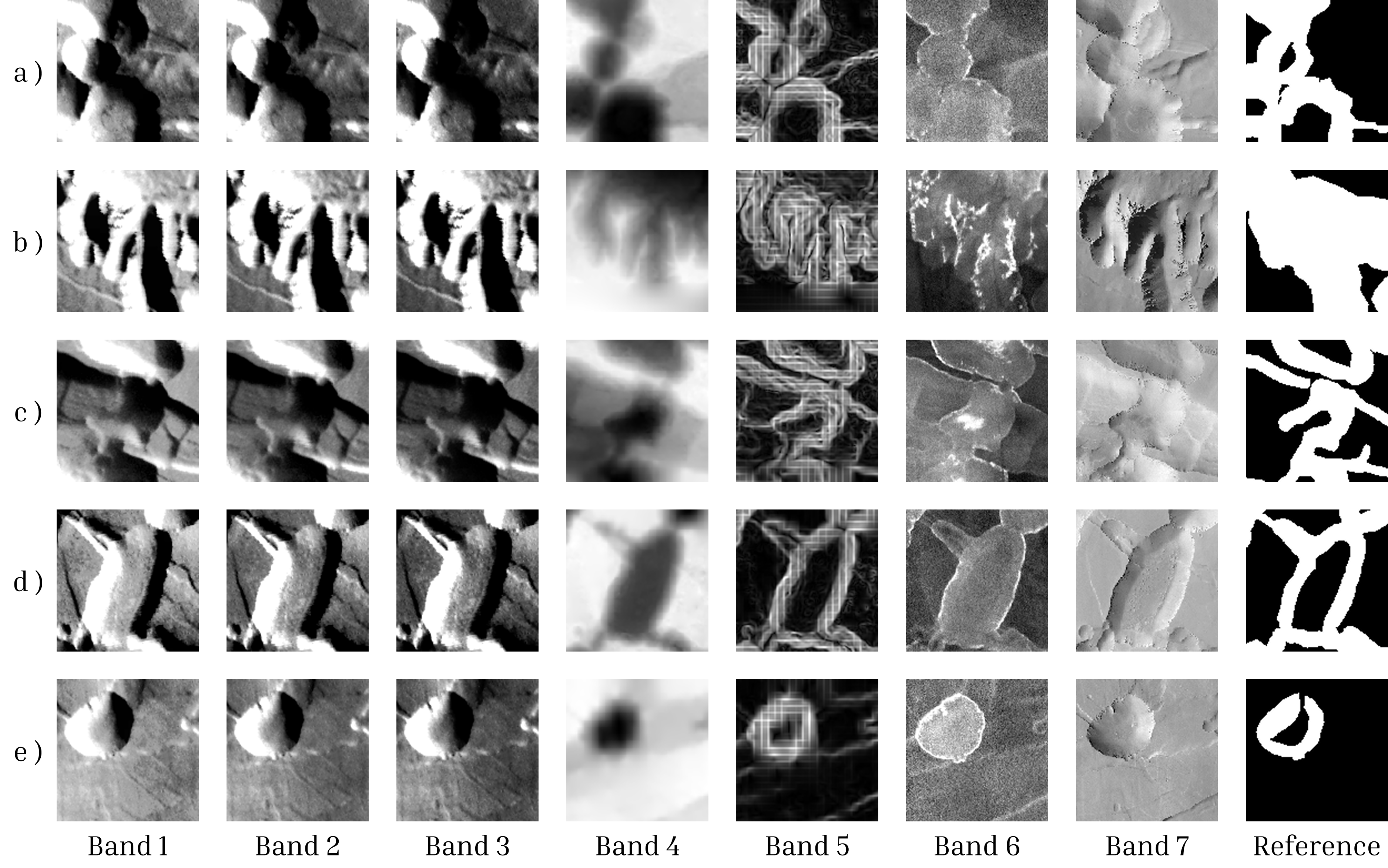}
    \caption{Representative scenes from the MMLSv2 isolated test. Compared to the baseline split, these examples highlight noticeable shifts in spatial context, texture, and landslide morphology, including fragmented multi component failures (row a), large continuous regions (row b), discontinuous and irregular patterns (row c), elongated and curved structures (row d), and small isolated events (row e). This shows the out-of-distribution nature of the isolated test set and its role as a challenging benchmark for evaluating model generalization beyond standard in-distribution settings. Band order: (1) Red, (2) Green, (3) Blue, (4) DEM, (5) Slope, (6) Thermal inertia, (7) Grayscale.}
    \label{fig:samples_isolated}
\end{figure*}

Figs. \ref{fig:mmlsv2_samples} and \ref{fig:samples_isolated} illustrate the variability captured by MMLSv2 and its isolated test split. The baseline split (Fig. \ref{fig:mmlsv2_samples}) shows substantial heterogeneity in landslide appearance, including compact failures, elongated and curved structures, and extended continuous regions across diverse geomorphological contexts. The isolated test set (Fig. \ref{fig:samples_isolated}) further amplifies this variability by introducing scenes with stronger structural differences, irregular geometries, altered spatial organization, and sparser landslide signatures. As these samples deviate more clearly from the training distribution, the isolated split defines a more challenging evaluation setting, where segmentation performance relies less on memorized appearance patterns and more on generalization across distinct geomorphological contexts.

\section{Benchmark experiments}

\subsection{Experimental setup}

To evaluate the proposed dataset, different state of the art semantic segmentation models were considered, including U-Net, U-Net++ \citep{Zhouunetplusplus}, PSPNet \citep{8100143}, DeepLabV3 \citep{Chen2017RethinkingAC}, DeepLabV3+ \citep{Yukunv3plus}, and SegFormer \citep{alvarezsegformer}. All models were trained under a unified configuration to ensure a consistent comparison. Training employed the Adam optimizer with a learning rate of 0.001, cross entropy loss, and a step based learning rate scheduler reducing the rate by a factor of 0.1 every 30 epochs. Models were trained for 100 epochs with a batch size of 128 on a single NVIDIA A100 GPU with 40 GB memory, using all seven input bands and basic data augmentation via random flips and rotations. Performance was evaluated using precision, recall, F1 score, foreground and background IoU, and mean IoU, while inference and total training time were reported to assess computational efficiency. For each architecture, the model achieving the highest validation mIoU was selected and evaluated on both the standard and isolated test splits.

\subsection{Results and analysis}



Table \ref{tab:inference_all_models} reports the quantitative performance of the evaluated architectures on the MMLSv2 test split. All models converge properly and achieve stable performance, with mIoU values in the 0.81-0.83 range, indicating that the task remains non-trivial and still leaves room for improvement. Class-wise analysis shows that most performance degradation occurs in the foreground class, which is consistently more challenging than background regions. This behaviour indicates that, while MMLSv2 supports reliable training, it retains sufficient complexity to reveal limitations in current architectures and to motivate further methodological advances in landslide segmentation.

\begin{table*}[t]
\caption{Quantitative comparison of the evaluated models on the MMLSv2 test set. While the evaluated models learn effective representations and achieve reasonable performance, the results indicate that landslide segmentation in MMLSv2 remains challenging, leaving clear room for further methodological improvements. Models were trained using all available input bands.}\label{tab:inference_all_models}
\centering
\resizebox{0.6\linewidth}{!}{%
\begin{tabular}{m{2.4cm}m{1.4cm}m{1cm}m{1.46cm}m{1.2cm}m{1.2cm}m{1.2cm}m{1.5cm}m{1.5cm}}
\toprule
\textbf{Method} & \textbf{Precision$\uparrow$} & \textbf{Recall$\uparrow$} & \textbf{F1-score$\uparrow$} & \textbf{IoU$_{BG}\uparrow$} & \textbf{IoU$_{FG}\uparrow$} & \textbf{mIoU$\uparrow$}  & \textbf{Inference time {\footnotesize (s)}} & \textbf{Training time {\footnotesize (h)}}\\
\midrule
U-Net \citep{Ronnebergerunet} & 0.858 & 0.868 & 0.863 & 0.868 & 0.759 & 0.814 & 0.005 & 0.088\\
U-Net++ \citep{Zhouunetplusplus} & 0.864 & 0.879 & 0.871 & 0.875 & 0.772 & 0.823 & 0.011 & 0.085\\
PSPNet \citep{8100143} & 0.866 & 0.884 & 0.875 & 0.878 & 0.778 & 0.828 & 0.005 & 0.027\\
DeepLabV3 \citep{Chen2017RethinkingAC} & 0.870 & 0.860 & 0.865 & 0.872 & 0.763 & 0.817 & 0.007 & 0.063\\
DeepLabV3+ \citep{Yukunv3plus} & 0.863 & 0.889 & 0.876 & 0.878 & 0.779 & 0.829 & 0.007 & 0.048\\
SegFormer \citep{alvarezsegformer} & 0.859 & 0.863 & 0.861 & 0.867 & 0.756 & 0.812 & 0.041 & 0.131\\
\bottomrule
\end{tabular}}
\vspace{1mm}
\\\footnotesize{\textit{\footnotesize *Inference time refers to the average latency per image.}}
\end{table*}

\begin{figure*}[t]
    \centering
    \includegraphics[width=0.76\linewidth]{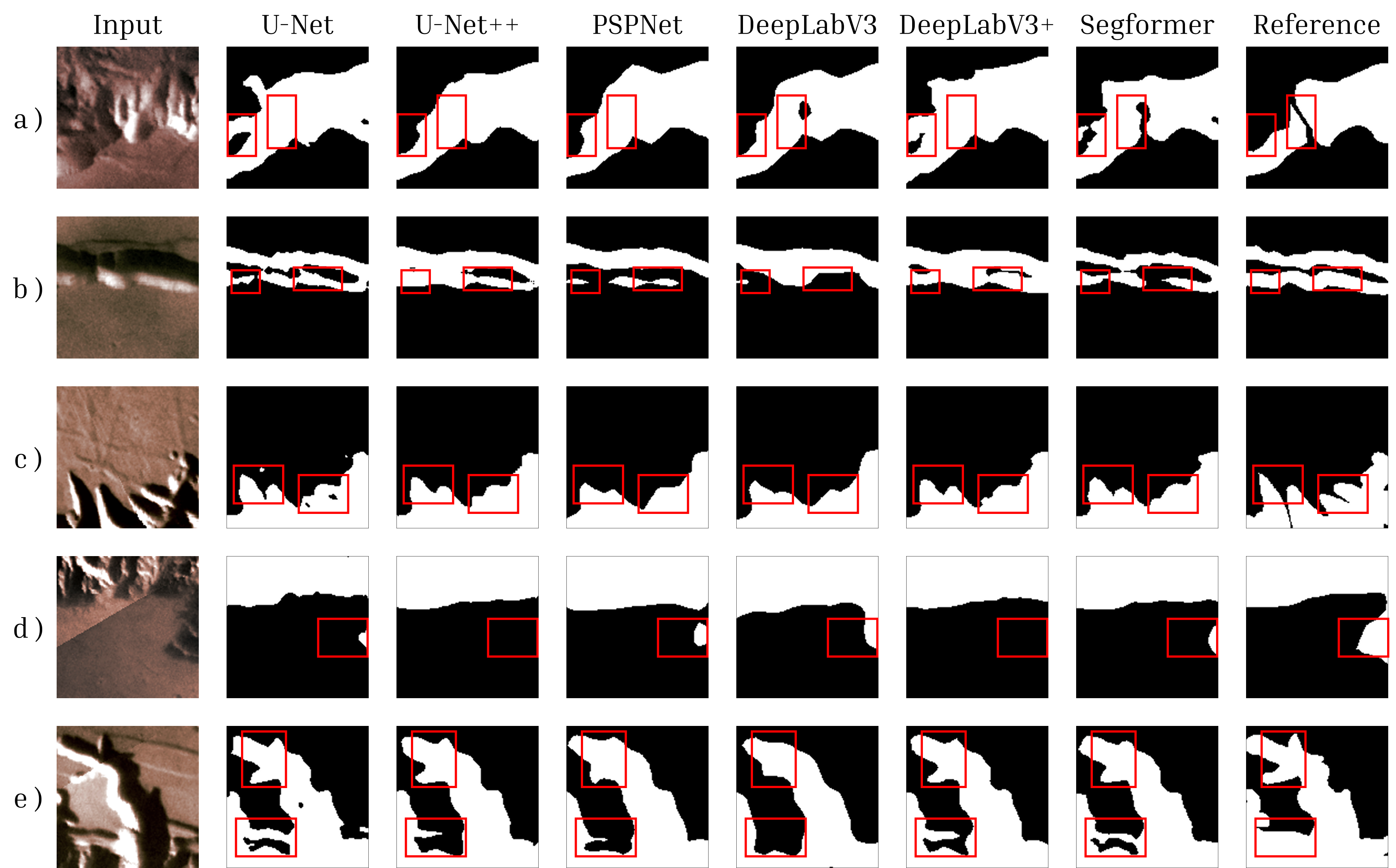}
    \caption{Qualitative comparison of landslide segmentation results obtained with the evaluated models trained using the full set of input bands in MMLSv2. While all models capture the main landslide structures, the highlighted regions reveal persistent errors in boundary delineation, small-scale failures, and fragmented or ambiguous areas, indicating room for improvement.}
    \label{fig:inference_all_models}
\end{figure*}

Fig. \ref{fig:inference_all_models} provides a qualitative complement to the quantitative results. All architectures recover the main landslide structures, particularly when failures form large, continuous regions, as shown in rows (a), (c), and (d). Their limitations become apparent for fragmented or weakly contrasted landslides. In rows such as (b) and (e), where failures appear as thin, discontinuous, or spatially sparse patterns, all models show partial detections, boundary inaccuracies, or missed regions. Even for well-defined landslide bodies, small-scale details and elongated structures are not consistently distinguished across predictions.

\begin{figure}[t]
    \centering
    \includegraphics[width=0.6\linewidth]{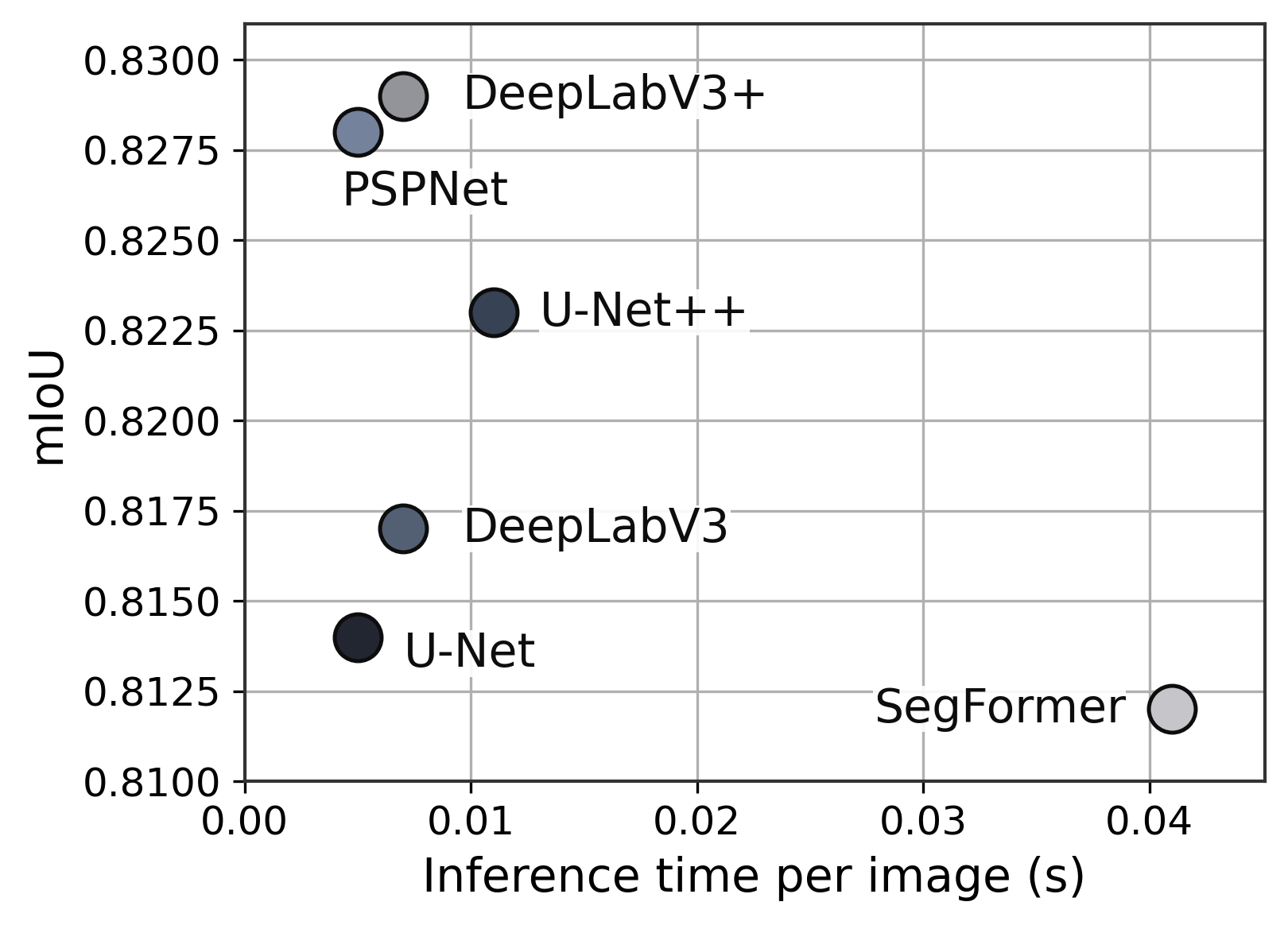}
    \caption{Trade-off between inference time per image and mIoU on the MMLSv2 test set for the evaluated architectures. Latency is measured on a single NVIDIA A100 40GB GPU.}
    \label{fig:trade_off_pallete}
\end{figure}

Regarding efficiency, training times follow architectural complexity. PSPNet is the fastest to train (0.027 h), followed by the U-Net variants, while SegFormer exhibits a substantially higher training cost (0.131 h), reflecting the overhead of its Transformer-based design. This is consistent with the inference-time trade-off shown in Fig. \ref{fig:trade_off_pallete}, where SegFormer incurs the highest latency without a corresponding gain in mIoU. In contrast, convolutional models cluster in a more favourable accuracy-efficiency region, achieving comparable performance at lower computational cost.

Table \ref{tab:inference_isolated} reports model performance on the isolated test set. Relative to the baseline split, all models exhibit a consistent mIoU drop, decreasing from above 0.80 to the 0.70-0.73 range. The degradation is most pronounced in foreground IoU, which falls from values above 0.70 to slightly above 0.60 in the best-performing model (SegFormer), while background performance remains stable. This shows the increased structural and spatial variability of landslide patterns in the isolated test set, confirming it as a more challenging benchmark for spatial generalization.

\begin{table}[t]
\caption{Quantitative performance of the evaluated models on the isolated test set of MMLSv2. Although all methods achieve reasonable performance, scores are consistently lower than on the baseline test split, reflecting the increased difficulty of the isolated setting. This mirrors realistic deployment conditions, where models trained on specific regions must generalize to unseen areas.}\label{tab:inference_isolated}
\centering
\resizebox{\columnwidth}{!}{%
\begin{tabular}{m{2.4cm}m{1.4cm}m{1cm}m{1.46cm}m{1.2cm}m{1.2cm}m{1.2cm}m{1.5cm}m{1.5cm}}
\toprule
\textbf{Method} & \textbf{Precision$\uparrow$} & \textbf{Recall$\uparrow$} & \textbf{F1-score$\uparrow$} & \textbf{IoU$_{BG}\uparrow$} & \textbf{IoU$_{FG}\uparrow$} & \textbf{mIoU$\uparrow$}  & \textbf{Inference time {\footnotesize (s)}}\\
\midrule
U-Net \citep{Ronnebergerunet} & 0.676 & 0.828 & 0.744 & 0.848 & 0.593 & 0.721 & 0.007\\
U-Net++ \citep{Zhouunetplusplus} & 0.701 & 0.743 & 0.722 & 0.851 & 0.564 & 0.708 & 0.020\\
PSPNet \citep{8100143} & 0.679 & 0.786 & 0.729 & 0.846 & 0.573 & 0.709 & 0.008\\
DeepLabV3 \citep{Chen2017RethinkingAC} & 0.727 & 0.754 & 0.740 & 0.862 & 0.588 & 0.725 & 0.016\\
DeepLabV3+ \citep{Yukunv3plus} & 0.699 & 0.714 & 0.706 & 0.847 & 0.546 & 0.696 & 0.014\\
SegFormer \citep{alvarezsegformer} & 0.684 & 0.856 & 0.761 & 0.855 & 0.614 & 0.735 & 0.080\\
\bottomrule
\end{tabular}}
\vspace{1mm}
\\\footnotesize{\textit{\footnotesize *Inference time refers to the average latency per image.}}
\end{table}



\begin{figure*}[t]
    \centering
    \includegraphics[width=0.76\linewidth]{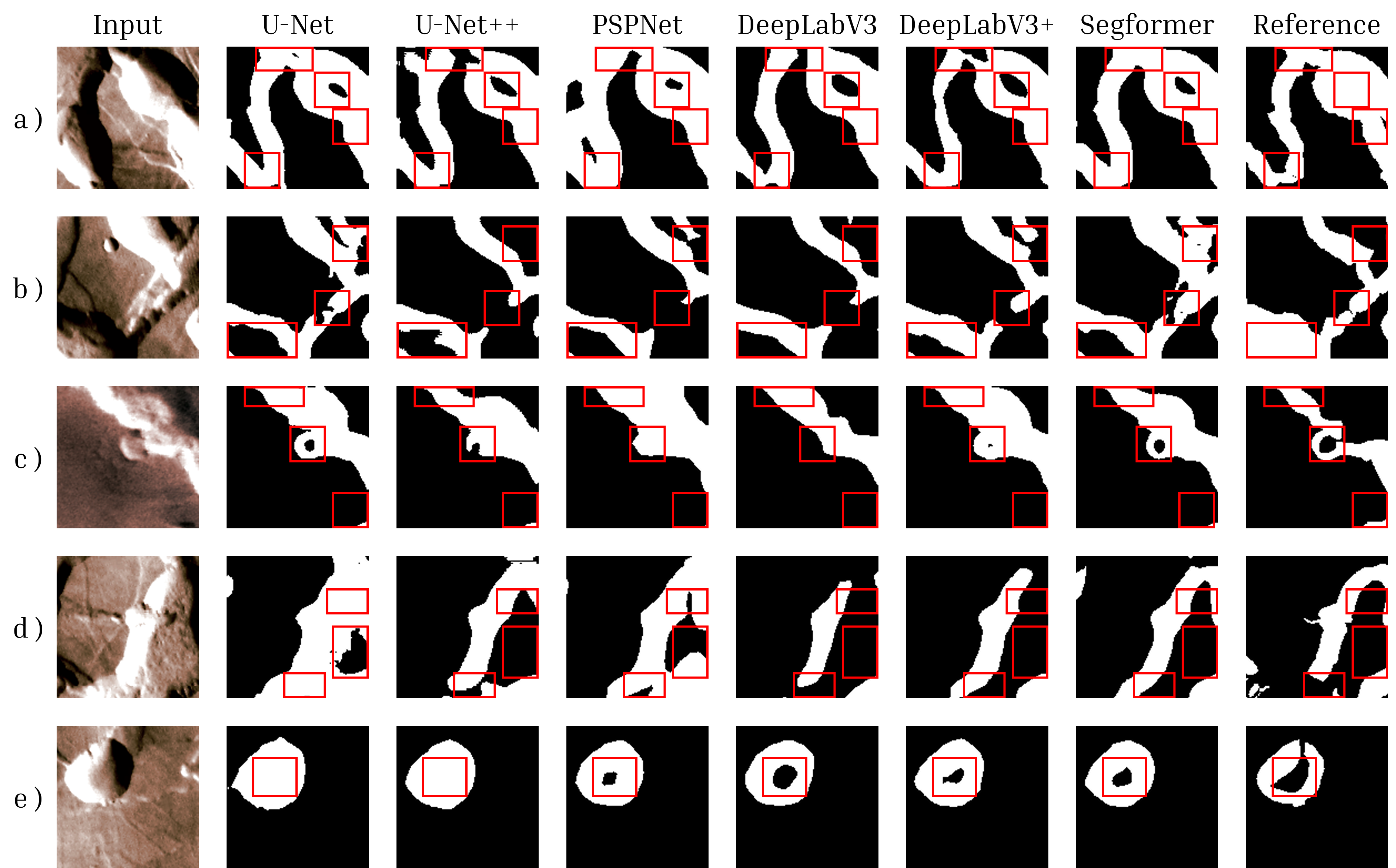}
    \caption{Qualitative comparison of landslide segmentation results on the isolated test set of MMLSv2 using the evaluated models trained with the full input configuration. 
    Red boxes indicate representative areas where predictions diverge from the reference masks.}
    \label{fig:inference_isolated}
\end{figure*}

Fig. \ref{fig:inference_isolated} confirms the trends observed in the quantitative evaluation of the isolated test set. Errors are frequent in fragmented and discontinuous regions, as well as in elongated or thin landslide structures, where predictions often break apart or miss affected areas, as shown in rows (a), (b), and (d). Even in more regular cases, such as rows (c) and (e), inaccuracies persist along irregular boundaries and narrow regions. This indicates that the isolated test set introduces morphological variability and spatial configurations insufficiently represented during training.

Next, Table \ref{tab:inference_band_combinations} shows the impact of different band combinations. Incorporating additional spectral and auxiliary bands yields a consistent mIoU improvement, increasing from 0.711 with RGB alone to 0.814 when using all seven bands, indicating that the added modalities provide complementary rather than redundant information. In particular, DEM and thermal cues steadily enhance foreground delineation, confirming their relevance for landslide characterization. Latency remains unchanged across configurations, while training time increases only marginally, showing that the performance gains are achieved without significant computational overhead.

\begin{table}[ht]
\caption{Impact of different band combinations using U-Net \citep{Ronnebergerunet} on the MMLSv2 test set. Segmentation performance improves consistently as additional spectral and auxiliary bands are incorporated, while the non-uniform gains across combinations indicate complementary rather than redundant modality contributions.}\label{tab:inference_band_combinations}
\centering
\resizebox{0.95\linewidth}{!}{%
\begin{tabular}{m{4.7cm}m{1cm}m{1cm}m{1cm}m{1.5cm}m{1.5cm}}
\toprule
\textbf{Band combination} & \textbf{IoU$_{BG}\uparrow$} & \textbf{IoU$_{FG}\uparrow$} & \textbf{mIoU$\uparrow$} & \textbf{Inference time {\footnotesize (s)}} & \textbf{Training time {\footnotesize (h)}}\\
\midrule
RGB & 0.808 & 0.614 & 0.711 & 0.005 & 0.082\\
RGB-DEM & 0.828 & 0.696 & 0.762 & 0.005 & 0.083\\
RGB-DEM-Slope & 0.865 & 0.745 & 0.805 & 0.005 & 0.083\\
RGB-DEM-Slope-Thermal & 0.870 & 0.753 & 0.812 & 0.005 & 0.085\\
RGB-DEM-Slope-Thermal-Gray & 0.868 & 0.759 & 0.814 & 0.005 & 0.088\\
\bottomrule
\end{tabular}}
\vspace{1mm}
\\\footnotesize{\textit{\scriptsize *Inference time refers to the average latency per image.}}
\end{table}

\begin{figure}[t]
    \centering
    \includegraphics[width=1\columnwidth]{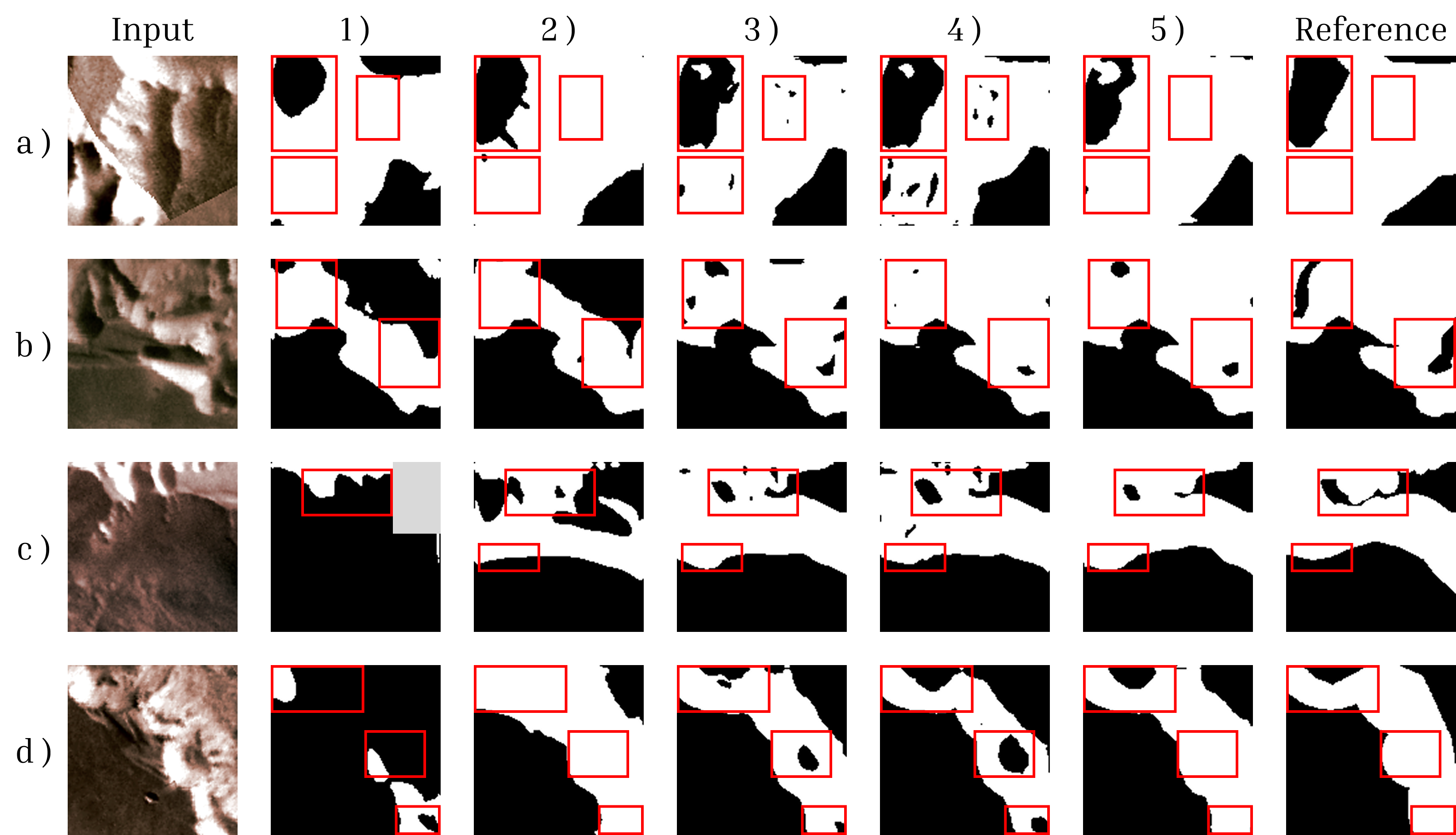}
    \caption{Inference results obtained with a U-Net model using different input band combinations on the MMLSv2 dataset. Each row shows an example image, while columns correspond to different band configurations: (1) RGB, (2) RGB-DEM, (3) RGB-DEM-Slope, (4) RGB-DEM-Slope-Thermal, and (5) RGB-DEM-Slope-Thermal-Grayscale. 
    }
    \label{fig:inference_band_combinations}
\end{figure}

Finally, Fig. \ref{fig:inference_band_combinations} presents qualitative segmentation results for different input band combinations. Consistent with the quantitative analysis, adding spectral and auxiliary bands progressively improves segmentation quality, reducing missed regions and producing more coherent landslide boundaries, particularly in fragmented or low-contrast areas. The full multispectral configuration yields the most stable delineations, with fewer false negatives and improved continuity of elongated and irregular structures.

\section{Limitations and future work}

Although we provide a broad evaluation of MMLSv2, some limitations should be acknowledged. The experimental setup is intentionally kept simple across models, without architecture-specific tuning, so the reported results should be interpreted as reference baselines rather than performance upper bounds, leaving room for future exploration of more specialized training strategies. Also, multimodal information is integrated through direct band concatenation, without investigating advanced fusion mechanisms. Alternative designs such as multi-branch architectures or modality-aware fusion could better exploit the complementary spectral and auxiliary inputs. Finally, the evaluation was limited to a subset of widely used architectures, and extending the analysis to additional model families could further characterize the challenges and potential of MMLSv2.

\section{Conclusions}

We introduced MMLSv2, a multimodal dataset for landslide segmentation integrating seven data channels spanning optical, topographic, and thermal information over 664 images, together with an isolated test set of 276 spatially disjoint samples designed to explicitly assess generalization under distribution shifts. A comprehensive benchmark across multiple architectures shows that, while models converge reliably and achieve competitive performance on the baseline split, landslide segmentation on MMLSv2 remains challenging, particularly in fragmented and structurally complex regions. Performance consistently degrades on the isolated test set, which removes spatial overlap with the training data and introduces landslide patterns not represented in the standard split. 

{
    \small
    \bibliographystyle{ieeenat_fullname}
    \bibliography{main}
}

\end{document}